\documentclass{article}

\PassOptionsToPackage{numbers}{natbib}
\usepackage[preprint]{neurips_2026}
\usepackage{amsmath}
\usepackage{multirow}
\usepackage{algorithm}
\usepackage{algorithmic}

\usepackage[utf8]{inputenc} 
\usepackage[T1]{fontenc}    
\usepackage{caption}
\usepackage{hyperref}       
\hypersetup{hidelinks}
\usepackage{booktabs}       
\usepackage{amsfonts}       
\usepackage{nicefrac}       
\usepackage{microtype}      
\usepackage{multirow}
\usepackage{xcolor}         
\usepackage{graphicx}

\title{Asymmetric Hierarchical Anchoring for Robust Audio–Visual Cross-Modal Generalization}

\author{%
  \begin{tabular}{ccc}
    Bixing Wu\textsuperscript{1,*} &
    Yuhong Zhao\textsuperscript{1,*} &
    Zongli Ye\textsuperscript{1,3,*} \\[2pt]
    Jiachen Lian\textsuperscript{2,\textdagger} &
    Xiangyu Yue\textsuperscript{3} &
    Gopala Anumanchipalli\textsuperscript{2}
  \end{tabular}\\[6pt]
  \normalfont
  \textsuperscript{1}Zhejiang University, China
  \qquad
  \textsuperscript{2}University of California, Berkeley, USA\\
  \textsuperscript{3}MMLab, Chinese University of Hong Kong, China\\[4pt]
  \texttt{bixingwu@zju.edu.cn}, \texttt{jiachenlian@berkeley.edu}\\
  \textsuperscript{*}Equal contribution.
  \qquad
  \textsuperscript{\textdagger}Corresponding author.
}

\begin{document}

\maketitle

\begin{abstract}
Audio--visual joint representation learning under Cross-Modal Generalization (CMG) aims to transfer knowledge from a labeled source modality to an unlabeled target modality through a unified discrete representation space. Existing symmetric frameworks often suffer from information allocation ambiguity, where the absence of structural inductive bias leads to semantic--specific leakage across modalities. We propose Asymmetric Hierarchical Anchoring (AHA), which enforces directional information allocation by designating a structured semantic anchor within a shared hierarchy. In our instantiation, we exploit the hierarchical discrete representations induced by audio Residual Vector Quantization (RVQ) to guide video feature distillation into a shared semantic space. To ensure representational purity, we replace fragile mutual information estimators with a GRL-based adversarial decoupler that explicitly suppresses semantic leakage in modality-specific branches, and introduce Local Sliding Alignment (LSA) to encourage fine-grained temporal alignment across modalities. Extensive experiments on AVE and AVVP benchmarks demonstrate that AHA consistently outperforms symmetric baselines in cross-modal transfer. Additional analyses on talking-face disentanglement experiment further validate that the learned representations exhibit improved semantic consistency and disentanglement, indicating the broader applicability of the proposed framework.
\end{abstract}

\section{Introduction}

Multimodal learning aims to mimic human perceptual integration, driving significant progress in tasks such as Visual Question Answering ~\cite{antol2015vqa,lei2018tvqa,li2022learning} and Audio--Visual Event Localization~\cite{AVE,xu2020cross,geng2025uniav}. In this work, we focus on audio--visual joint representation learning, where early research predominantly employed implicit continuous representations (e.g., CLIP~\cite{radford2021learning}, ImageBind~\cite{girdhar2023imagebind}) to align semantics via contrastive learning to reduce the distance between different modalities in high-dimensional semantic space, the field has increasingly shifted towards explicit discrete representations based on codebooks (e.g.,~\cite{codis,cmcm,lu2022unified,chen2025semhitok}) or prototypes~\cite{snell2017prototypical,chen2023protoclip} to represent different modalities. As shown in \autoref{fig:cmgtask} (2), to mitigate the Modality Gap, a modality-specific branch was introduced, which enables discrete representations to capture cross-modal semantic information. The use of discrete space enables aggregation of similar input features in high-dimensional space, allowing complex feature representations to be achieved with a small number of latent codes. However, existing discretization methods often rely on the idealized assumption of perfect alignment, failing to address the semantic gaps and annotation cost disparities inherent in unconstrained videos. To overcome these limitations, Cross-Modal Generalization (CMG) task and the Unicode framework~\cite{unicode} were introduced. This approach achieves fine-grained unified discrete representations on unlabeled paired data, enabling zero-shot transfer capabilities under single-modality training.

\begin{figure}[H]
  \centering
  \includegraphics[width=\columnwidth]{images/task.pdf}
  \caption{(1) \textbf{The Cross-Modal Generalization (CMG) Task Definition}; (2) \textbf{Illustration of the challenges and goals in audio-visual joint representation learning}. (a) \textit{Modality Gap}: Audio (\textcolor{blue}{blue}) and video (\textcolor{red}{red}) features are separated in the latent space due to modality-specific biases. This is why we need to separate the semantic and specific parts; (b) \textit{Bad Unified Representation}: A naive unification leads to information leakage and semantic codebook collapse, where semantic and modality-specific factors are entangled around discrete codes; (c) \textit{Good Unified Representation}: A well-structured unified space aligns cross-modal semantic features around shared discrete representations while isolating modality-specific variations.}
  \label{fig:cmgtask}
\end{figure}

Although the current explicit representations~\cite{codis,cmcm,lu2022unified,chen2025semhitok} achieve certain effects on standard downstream tasks, our analysis and experiments show that this does not necessarily mean effective disentanglement of semantic and specific information. In experiments we designed, we find that due to the weak information retention capacity of the previous discrete semantic codebook, the shared space of the symmetric structure undergoes collapse, that is, semantic information tends to flow more towards the specific branches which have less constraints rather than towards the discrete unified representation desired by the CMG task. This exposes potential limitations of the symmetric structure in disentanglement, which we define as Information Allocation Ambiguity. These works~\cite{unicode,fcid} often use Mutual Information (MI) estimators, such as the Contrastive Log-ratio Upper Bound (CLUB)~\cite{cheng2020club}, to minimize the correlation between the two streams. However, we find that MI minimization via variational upper-bound estimators (e.g., CLUB), which relies on learning a conditional model, becomes unreliable in unconstrained high-dimensional settings.

Our work attempts to answer a more structured question: in audio--visual joint representation, whether semantic information is explicitly concentrated in a designated shared subspace and does not leak into modality-specific branches. This distinction is particularly critical in generative or controllable modeling tasks. To address this, we propose an Asymmetric Hierarchical Anchoring (AHA) structure. Motivated by the fact that audio is commonly used as a conditioning signal in generative settings and often provides a compact and robust cue for high-level semantics under unconstrained visual noise~\cite{aytar2016soundnet,narasimhan2022strumming, biner2024sonicdiffusion, cheng2024dawn, gao2025wan}, we anchor semantics on the audio modality, utilizing a single-branch Residual Vector Quantization (RVQ)~\cite{lee2022autoregressive} structure to construct an asymmetric semantic anchor point. RVQ has been widely adopted in generation and representation learning(especially in audio field)~\cite{defossez2022high,kim2024efficient,zhang2025mbcodec,kim2025improving} due to its ability to naturally decompose a signal into a coarse-to-fine hierarchy where the primary quantization layers capture the most salient semantic concepts. We designate this layer as a semantic anchor, forcing the video's semantic encoder to align with this predefined discrete target. This provides a structural prior that guides visual feature distillation, effectively resolving the allocation dilemma. On this basis, we introduce a more effective Gradient Reversal Layer (GRL)-based Adversarial Decoupler. By treating disentanglement as a min-max game, we suppress modality-specific information from the video semantic branch. Finally, we introduce Local Sliding Alignment (LSA) to refine cross-modal alignment at a granular temporal level for unconstrained sequences. This entire structure has been verified as effective through a series of our experiments. Our main contributions are summarized as follows:
\begin{itemize}
    \item We introduce an Asymmetric architecture that leverages the hierarchical nature of audio RVQ as a structural anchor to provide a stable target for visual semantic distillation.
    \item We propose a robust GRL-based Adversarial Decoupler that outperforms traditional variational upper-bound MI estimators in separating semantics from modality-specific information.
    \item We propose AHA, a novel Cross-Modal Generalization framework that achieves effective semantic-specific disentanglement; the quality of the learned representations is further validated by a talking-face diagnostic experiment, suggesting broader applicability beyond classification tasks.
\end{itemize}

\section{Related Works}

\textbf{Multimodal Unified Representations}: Recent research on multimodal unified representation has primarily focused on bridging the semantic gap among heterogeneous data. Mainstream approaches typically align modalities by projecting them into a shared latent space~\cite{petridis2018audio,andonian2022robust,sarkar2024xkd} or employing modal-general encoders for unified cross-modal feature extraction~\cite{chen2020uniter,wang2022vlmixer}. To further facilitate cross-modal synergy, cross-modal knowledge distillation is widely adopted for implicit information transfer~\cite{sarkar2024xkd,huo2024c2kd,jeong2025multi}, while representation bridging techniques enhance complementarity by connecting distinct continuous spaces~\cite{wang2023connecting,grassucci2025closing}. Furthermore, to enhance interpretability and structure, discretization methods based on codebooks or prototypes have gained increasing attention, mapping continuous features into compact discrete forms~\cite{lu2022unified,jin2023unified,yu2024image}.

\textbf{Mutual Information and GRL Disentanglement}: In self-supervised learning, mutual information is commonly used for cross-modal alignment, such as in InfoNCE~\cite{oord2018representation}, Contrastive Predictive Coding (CPC)~\cite{oord2018representation}, and MINE~\cite{belghazi2018mutual}, which typically implicitly maximize the mutual information lower bound through contrastive learning or neural estimators. In contrast, for disentanglement between information, variational upper-bound MI estimators like CLUB~\cite{cheng2020club} have difficulty in fitting complex, high-dimensional conditional distributions in unconstrained scenarios. However, in the asymmetric anchor structure we introduced, adversarial learning via Gradient Reversal Layers (GRL)~\cite{ganin2015unsupervised} benefits from the anchor, offering a more straightforward information stripping paradigm that provides superior robustness in identifying and eliminating biases without complex density estimation and has been applied in many applications~\cite{zhang2018mitigating,yu2020learning,zhang2023towards,ju2024naturalspeech}. Consequently, we adapt this adversarial paradigm to purge modality-specific statistics, ensuring a purer semantic space for cross-modal transfer.

\section{Method}

\subsection{Overview and Notations}
\label{sec:overview}

As illustrated in \autoref{fig:pipeline}, our framework aligns cross-modal semantics through a dual-stream architecture comprising audio and video branches. Let $x_a$ and $x_v$ denote the input audio and video sequences, respectively. The architecture utilizes three primary encoders: \textbf{Audio Encoder ($E_a$)}: Extracts features from the audio input, denoted as the audio backbone output. \textbf{Video Semantic Encoder ($E_{v\_sem}$)}: Extracts the continuous semantic feature sequence $Z_{sem}^V \in \mathbb{R}^{T \times D}$, which captures high-level content shared across modalities. \textbf{Video Specific Encoder ($E_{v\_spec}$)}: Extracts the modality-exclusive feature sequence $Z_{spec}^V$, capturing video-specific attributes.

The core of our alignment strategy is a hierarchical Residual Vector Quantization (RVQ) module containing $n$ layers which is divided into the Primary Codebook (shared semantic anchor) and Residual Specific Layers. 
After quantization, we obtain the discrete representations \(\mathcal{A}_{unit} = \{A_1, \dots, A_T\}\) for audio and \(\mathcal{V}_{unit} = \{V_1, \dots, V_T\}\) for video.
Finally, the Audio Decoder ($G_a$) and Video Decoder ($G_v$) reconstruct the original signals from these disentangled representations.

\subsection{Structure of Asymmetric Hierarchical Anchoring}
\label{sec:rvq}

To align cross-modal semantics, we propose Asymmetric Hierarchical Anchoring (AHA), which leverages RVQ's hierarchical structure to impose directional semantic constraints, enabling consistent alignment without symmetric decomposition.

\textbf{Shared Semantic Alignment (Layers $1 \sim k$)}: 
Due to RVQ's tendency to aggregate semantic information in lower layers, the first $k$ layers serve as a cross-modal Semantic Anchor.  
For the audio branch, the input is processed by these shared layers as
\begin{equation}
\mathbf{q}_l = \operatorname{Quantizer}(\mathbf{r}_{l-1}, \mathcal{C}_{l} \subset \mathcal{C}_{\mathrm{shared}}); \quad l \le k.
\end{equation}
where $\mathcal{C}_{l}$ is the shared codebook at layer $l$, and $\mathbf{r}_{l-1}$ is the residual from the previous layer (or the input for $l=1$).

For the video branch, the continuous semantic feature $Z_{sem}^V$ is also projected onto this shared codebook space. This constraint forces the discretized video representation $\mathcal{V}_{unit}$ to capture only the high-level semantic content that is align with audio semantic and common to both modalities.

\textbf{Asymmetric Refinement (Layers $k+1 \sim n$)}: 
Under the constraint of reconstruction loss, the audio branch extends beyond the shared layers to guide the modality-specific information flow to higher codebooks. The final inputs to the decoders are formulated as:
\begin{equation}
    \mathcal{V}_{unit} = \sum_{l=1}^{k} \mathbf{q}_l^V;
    \quad \\ 
    \mathcal{A}_{unit} = \underbrace{\sum_{l=1}^{k} \mathbf{q}_l^A}_{\text{Semantic Anchor}} + \underbrace{\sum_{l=k+1}^{n} \mathbf{q}_l^A}_{\text{Specific features}}
\end{equation}

\subsection{Adversarial Disentanglement via Contrastive Learning}
\label{adv_grl}

\begin{figure*}[t]
  \centering
  \includegraphics[width=\textwidth]{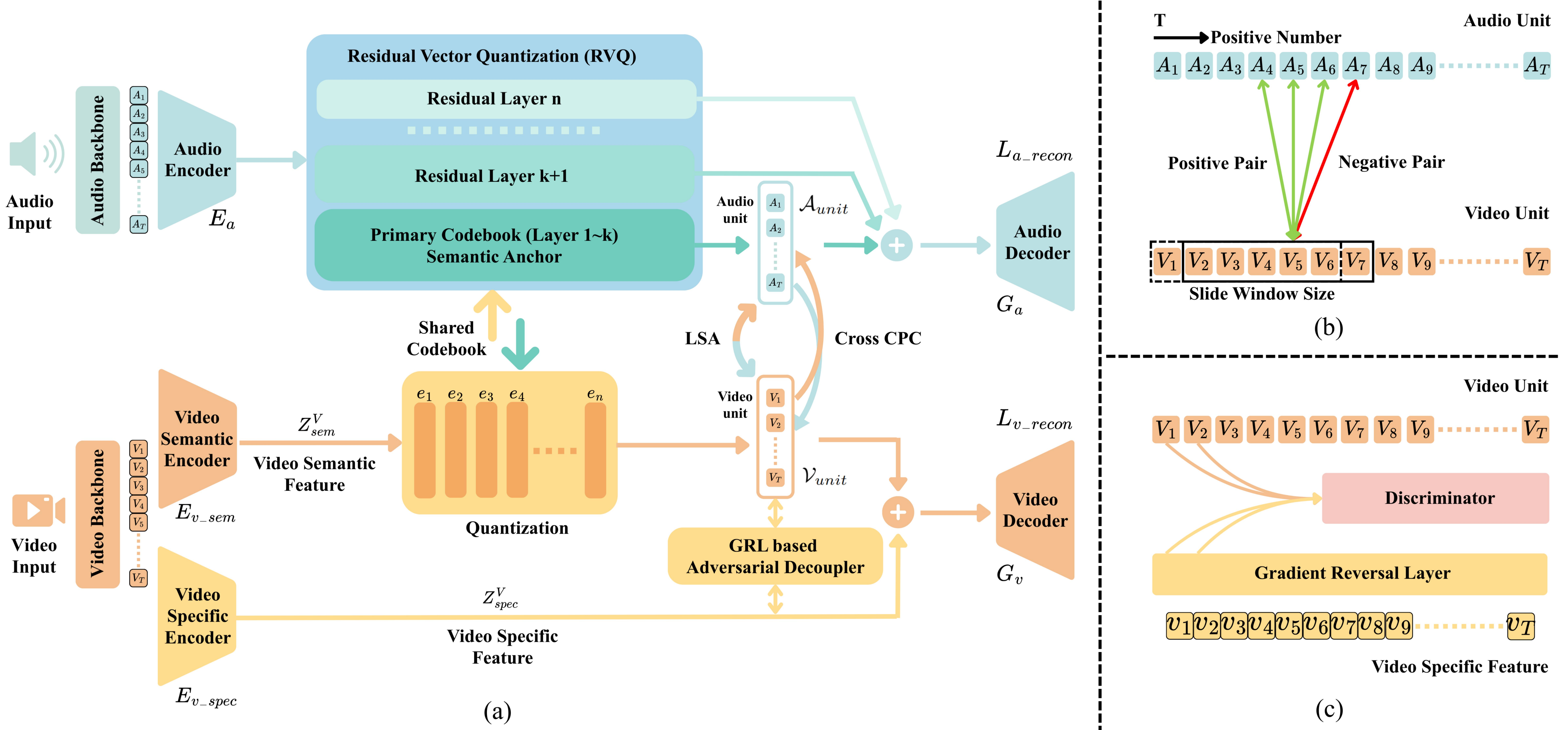}
  \caption{\textbf{Overview of the proposed AHA architecture}. (a) \textit{The main pipeline of our entire model}. Through our experiments, we found that the effective semantic information in the audio RVQ is generally concentrated in the first two layers. Therefore, we make slight adjustments to $k$ according to the information density required by the task. For the CMG task, to unify the comparison with previous state-of-the-art work, we set $k=1$ by default. For the subsequent Talking Face Disentanglement Experiments~\autoref{fig:face}, we set $k=2$; (b) \textit{Framework of Local Sliding Alignment (LSA)}; (c) \textit{Framework of Gradient Reversal Layer(GRL)-based Adversarial Decoupler}. We also discuss the choice of $k$ and LSA hyperparameters in~\autoref{MoreAblation}.}
  \label{fig:pipeline}
\end{figure*}

In this section, we elaborate on the adversarial mechanism used to ensure orthogonality between the semantic and specific representations defined in \autoref{sec:overview}.
Specifically, we aim to ensure that the specific feature $Z_{spec}^V$ extracted by $E_{v\_spec}$ remains independent of the semantic content encapsulated in the Video Units $\mathcal{V}_{unit}$. To achieve this, we employ a Gradient Reversal Layer (GRL)-based Adversarial Decoupler shown in \autoref{fig:pipeline}(c).

\textbf{Adversarial Contrastive Objective}: 
Our objective is to decouple the specific features $v_t \in \mathcal{V}_{spec}$ and the semantic units $V_t \in \mathcal{V}_{unit}$. We formulate this as a min-max game involving $E_{v\_spec}$ and a conditional discriminator $D_\phi$.
The discriminator acts as a critic, attempting to distinguish the matched pair $(v_t, V_t)$ from negative samples. Conversely, $E_{v\_spec}$ aims to deceive the discriminator via the GRL $\mathcal{R}_\lambda(\cdot)$.
Let $s(\mathbf{x}, \mathbf{y}) = D_\phi(\mathcal{R}_\lambda(\mathbf{x}), \mathbf{y})$ denote the similarity score. The unified adversarial objective is:
\begin{equation}
\min\limits_{D_\phi} \max\limits_{E_{v\_spec}} \mathcal{L}_{grl} = - \mathbb{E} \left[ \log \frac{\exp(s(v_t, V_t) / \tau)}{\sum_{V' \in \{\mathcal{V}_{unit}\} \cup \mathcal{N}} \exp(s(v_t, V') / \tau)} \right]
\end{equation}
where $\tau$ is the temperature and $\mathcal{N}$ denotes the set of negative semantic units.

\textbf{Velocity-Aware Anchor Sampling}: 
We define the semantic velocity $\delta_t = \| V_{t+1} - V_t \|_2$. The probability of sampling a time step $t$ as an anchor is proportional to its velocity: $P_{sample}(t) = \frac{\delta_t}{\sum_{\tau} \delta_\tau + \epsilon}$
This ensures the discriminator focuses on dynamic semantic changes rather than static frames. For implementation details, please refer to \autoref{alg:grl_velocity}.

\subsection{Local Sliding Alignment}
To handle the inherent asynchrony between visual and acoustic signals in unconstrained scenarios, we propose a Local Sliding Alignment (LSA) mechanism. 
As illustrated in \autoref{fig:pipeline}(b), instead of enforcing strict global matching, LSA encourages local synchronization between Audio Units 
 $\mathcal{A}_{unit}$ and Video Units $\mathcal{V}_{unit}$. 
By explicitly modeling local alignment, LSA facilitates more robust cross-modal correspondence under real-world conditions.

We define a Local Search Scope $\Omega_t = \{ j \mid |t - j| \le R \}$ and a Positive Tolerance set $\mathcal{P}_t = \{ j \mid |t - j| \le n_{pos} \}$ (where $n_{pos} \le R$).
We construct a soft-target distribution $Y_{t,j}$ uniformly over $\mathcal{P}_t$. The predicted alignment probability $P_{t,j}$ is computed via a masked softmax over the audio and video units within $\Omega_t$:
\begin{equation}
\begin{split}
    Y_{t,j} = \frac{1}{|\mathcal{P}_t|} \mathbb{I}(j \in \mathcal{P}_t), \quad
    P_{t,j} = \frac{\exp(A_t^{\top} V_j / \tau)}{\sum_{m \in \Omega_t} \exp(A_t^{\top} V_m / \tau)}
\end{split}
\end{equation}

The alignment loss minimizes the bidirectional cross-entropy:
\begin{equation}
\begin{split}
    \mathcal{L}_{align} = - \frac{1}{2T} \sum_{t=1}^{T} \bigg( & \sum_{j \in \Omega_t} Y_{t,j} \log P_{t,j}^{A \to V} + \sum_{j \in \Omega_t} Y_{t,j} \log P_{t,j}^{V \to A} \bigg).
\end{split}
\end{equation}

\subsection{Fine-Grained Alignment via Cross-CPC and MM-EMA}
To bridge the modality gap at a fine-grained level, we employ Cross-Modal Contrastive Predictive Coding (Cross-CPC). We use modality-specific LSTMs to aggregate historical contexts from the quantized units, denoted as $\mathbf{h}_t^V$ and $\mathbf{h}_t^A$. Crucially, the context of one modality predicts the future semantic units of the counterpart (e.g., $\mathbf{h}_t^V$ predicts $A_{t+step}$). The symmetric InfoNCE objective is:
\begin{equation}
\mathcal{L}_{CPC} = \frac{1}{2} \left( \mathcal{L}_{V \to A} + \mathcal{L}_{A \to V} \right),\quad 
\mathcal{L}_{V \to A} \propto - \sum_{t, step} \log \frac{\exp(A_{t+step}^{\top} W_{step}^V \mathbf{h}_t^V)}{\sum_{A_j \in \mathcal{N}} \exp(A_j^{\top} W_{step}^V \mathbf{h}_t^V)}.
\end{equation}

To encourage feature coupling in the shared latent space, we adopt Multi-Modal Exponential Moving Average (MM-EMA) for codebook updates. 
Unlike standard EMA, which updates each code vector based on a single modality, MM-EMA aggregates statistics from both audio and video features to ensure that each codebook vector \(\mathbf{e}_i\) represents the centroid of the joint distribution.

Specifically, for code \(i\), we maintain a moving average of cluster size \(N_i\) and embedding sum \(\mathbf{m}_i\), updated as:
\begin{equation}
\begin{split}
N_i^{(t)} &= \gamma N_i^{(t-1)} + (1-\gamma)(n_i^a + n_i^b), \\
\mathbf{m}_i^{(t)} = \gamma \mathbf{m}_i^{(t-1)} + &(1-\gamma) \left[ \sum_{j=1}^{n_i^a} A_{i,j} + \sum_{j=1}^{n_i^b} V_{i,j} \right],\quad
\mathbf{e}_i^{(t)} = \mathbf{m}_i^{(t)} / N_i^{(t)},
\end{split}
\end{equation}
where \(n_i^a, n_i^b\) denote the number of features assigned to code \(i\) from each modality, and \(\gamma\) is the decay factor. 

To further align encoder outputs with these shared centroids, we modify the commitment loss to include both modalities:
\begin{equation}
    \mathcal{L}_{commit}^a = \beta \|A_i - \operatorname{sg}[\mathbf{e}_i^a]\|_2^2 + \frac{\beta}{2} \|A_i - \operatorname{sg}[\mathbf{e}_i^b]\|_2^2.
\end{equation}
The commitment loss for video $\mathcal{L}_{commit}^v$ is defined in the same manner. This approach stabilizes quantization while pulling features from each modality closer to the shared semantic anchors.

\subsection{Total Loss Function}
The full objective function is a weighted combination of reconstruction, alignment, and disentanglement losses:
\begin{equation}
\begin{split}
    \mathcal{L}_{total} &= L_{a\_recon} + L_{v\_recon} + \mathcal{L}_{VQ}+ \mathcal{L}_{grl} + \mathcal{L}_{CPC} + \mathcal{L}_{align},
\end{split}
\end{equation}
where $L_{a\_recon}$ and $L_{v\_recon}$ correspond to the reconstruction losses shown in \autoref{fig:pipeline}(a), $\mathcal{L}_{VQ}$ sum of the commitment loss from MM-EMA and the commitment loss introduced by the additional audio RVQ layer. Each loss term targets a distinct component with naturally compatible gradient scales. We conducted sensitivity experiments and found performance remains robust across a range of weight combinations, with no single term dominating. 
\section{Experiment}

\subsection{Cross-Modality Downstream Tasks}
\subsubsection{Experimental Settings}
\textbf{Pre-training}: We follow Unicode's CMG evaluation protocol, select the same audio backbone and video backbone, pre-train on unlabeled paired audio--video data, freeze the encoders, train the same linear head using labeled source-modality features and evaluate zero-shot on the target-modality (V$\rightarrow$A / A$\rightarrow$V). Unless otherwise specified, we use the same single-layer linear classifier and settings as Unicode. Pre-training is conducted on VGGSound-AVEL 40K~\cite{vggsound2,vggsound1}, a VGGSound-derived in-the-wild dataset with naturally noisy audio--visual correspondence. Recent re-annotation of VGGSound reports incomplete labels, overlapping classes, and modality misalignment, with 48.43\% of original test samples containing incorrect labels or misaligned modalities and 40\% containing background music, voice-over, or static images~\cite{zverev2025vggsounder}. 
Thus, our CMG setting is not clean-pair pre-training but realistic noisy audio--visual learning. (Please see the \autoref{CMGTasks} for more implementation details). We also discuss training efficiency in \autoref{train_details}.

\textbf{Downstream Tasks}: We evaluate four standard CMG tasks, each in both transfer directions: \textbf{(i) AVE}~\cite{AVE}: cross-modal event classification on AVE; \textbf{(ii) AVVP}~\cite{AVVP}: cross-modal event localization on AVVP; \textbf{(iii) AVE $\rightarrow$ AVVP}: train on AVE classification and evaluate fine-grained localization on AVVP; \textbf{(iv) UCF(v) $\leftrightarrow$ VGG(a)}: cross-dataset cross-modal classification between UCF-101~\cite{soomro2012ucf101} (vision) and VGGSound (audio). (Details are shown in \autoref{CMGTasks}) We report accuracy for classification tasks and F1 for localization tasks.

\textbf{Baselines}: We compare with MST~\cite{mst}, CODIS~\cite{codis}, TURN~\cite{turn}, CMCM~\cite{cmcm}, and Unicode~\cite{unicode}. We also report DCID~\cite{fcid} and FCID~\cite{fcid} as reference results. They use additional text supervision (AVT) during unified-representation pre-training.

\begin{table}[htbp]
\centering
\small

\begin{minipage}[t]{0.47\textwidth}
  \centering
  \setlength{\tabcolsep}{2pt}
  \renewcommand{\arraystretch}{0.85}
  \captionof{table}{Cross-modal generalization on downstream CMG tasks.
    In the tables, V $\rightarrow$ A denotes training with video-modality
    labels and testing on audio, whereas A $\rightarrow$ V denotes the reverse.
    \textbf{Bold} indicates the best performance, and \textcolor{blue}{blue}
    denotes the improvement over \textcolor{red}{Unicode}.}
  \label{tab:cmg}
  \vspace{4pt}
  \resizebox{\textwidth}{!}{%
  \begin{tabular}{lccccccccc}
    \toprule
    \multirow{2}{*}{Method}
      & \multicolumn{2}{c}{AVE}
      & \multicolumn{2}{c}{AVVP}
      & \multicolumn{2}{c}{AVE$\rightarrow$AVVP}
      & \multicolumn{2}{c}{UCF(v)$\leftrightarrow$VGG(a)}
      & \multirow{2}{*}{Avg.} \\
    \cmidrule(lr){2-3}\cmidrule(lr){4-5}\cmidrule(lr){6-7}\cmidrule(lr){8-9}
      & V$\rightarrow$A & A$\rightarrow$V
      & V$\rightarrow$A & A$\rightarrow$V
      & V$\rightarrow$A & A$\rightarrow$V
      & V$\rightarrow$A & A$\rightarrow$V & \\
    \midrule
    MST(AV)
      & 19.5 & 23.1 & 22.7 & 24.5 & 29.5 & 36.4 & 45.7 & 43.1 & 30.56 \\
    CODIS(AV)
      & 36.8 & 39.7 & 32.7 & 32.6 & 40.8 & 40.6 & 50.8 & 45.2 & 39.90 \\
    TURN(AV)
      & 37.6 & 39.2 & 32.4 & 32.2 & 40.6 & 41.4 & 50.4 & 46.1 & 39.99 \\
    CMCM(AV)
      & 46.3 & 45.8 & 36.1 & 35.2 & 47.1 & 48.2 & 51.2 & 48.3 & 44.78 \\
    \textcolor{red}{Unicode}(AV)
      & 49.7 & 52.3 & 59.7 & 63.1 & 48.9 & 50.2 & 64.4 & 60.6 & 56.11 \\
    \midrule
    DCID(AVT)
      & 54.5 & 55.0 & 40.9 & 41.6 & 56.5 & 53.6 & 68.1 & 61.7 & 53.99 \\
    FCID(AVT)
      & 55.9 & 55.0 & 43.6 & 45.1
      & \textbf{57.4} & \textbf{58.5}
      & 69.6 & 62.0 & 55.89 \\
    \midrule
    \textbf{Ours}(AV)
      & \textbf{57.1} & \textbf{59.4}
      & \textbf{73.4} & \textbf{70.8}
      & 52.5 & 51.3
      & \textbf{70.3} & \textbf{63.1} & \textbf{62.24} \\
    \textbf{Improvements}
      & \textcolor{blue}{$+7.4$}  & \textcolor{blue}{$+7.1$}
      & \textcolor{blue}{$+13.7$} & \textcolor{blue}{$+7.7$}
      & \textcolor{blue}{$+3.6$}  & \textcolor{blue}{$+1.1$}
      & \textcolor{blue}{$+5.9$}  & \textcolor{blue}{$+2.5$}
      & \textcolor{blue}{$+6.13$} \\
    \bottomrule
  \end{tabular}}
\end{minipage}%
\hfill
\begin{minipage}[t]{0.52\textwidth}
  \centering

  \setlength{\tabcolsep}{2pt}
  \renewcommand{\arraystretch}{0.85}
  \captionof{table}{Ablation studies on different loss.}
  \label{tab:loss}
  \vspace{4pt}
  \resizebox{\textwidth}{!}{%
  \begin{tabular}{lccccccccc}
    \toprule
    \multirow{2}{*}{Method}
      & \multicolumn{2}{c}{AVE}
      & \multicolumn{2}{c}{AVVP}
      & \multicolumn{2}{c}{AVE$\rightarrow$AVVP}
      & \multicolumn{2}{c}{UCF(v)$\leftrightarrow$VGG(a)}
      & \multirow{2}{*}{Avg.} \\
    \cmidrule(lr){2-3}\cmidrule(lr){4-5}\cmidrule(lr){6-7}\cmidrule(lr){8-9}
      & V$\rightarrow$A & A$\rightarrow$V
      & V$\rightarrow$A & A$\rightarrow$V
      & V$\rightarrow$A & A$\rightarrow$V
      & V$\rightarrow$A & A$\rightarrow$V & \\
    \midrule
    \textbf{Ours}
      & \textbf{57.1} & \textbf{59.4}
      & \textbf{73.4} & \textbf{70.8}
      & \textbf{52.5} & \textbf{51.3}
      & \textbf{70.3} & 63.1
      & \textbf{62.24} \\
    w/o $L_{align}$
      & 55.4 & 58.9 & 65.6 & 68.1
      & 52.3 & 51.1 & 67.6 & \textbf{63.3} & 60.29 \\
    w/o $L_{grl}$
      & 47.3 & 50.2 & 58.1 & 62.9
      & 49.6 & 48.2 & 66.4 & 61.8 & 55.56 \\
    $L_{grl}\rightarrow L_{CLUB}$
      & 54.2 & 56.4 & 64.9 & 65.5
      & 52.4 & 50.8 & 65.1 & 62.7 & 59.00 \\
    \bottomrule
  \end{tabular}}

  \vspace{26pt} 

  \setlength{\tabcolsep}{2pt}
  \renewcommand{\arraystretch}{0.85}
  \captionof{table}{Ablation studies on different structures.}
  \label{tab:structure}
  \vspace{4pt}
  \resizebox{\textwidth}{!}{%
  \begin{tabular}{lccccccccc}
    \toprule
    \multirow{2}{*}{Method}
      & \multicolumn{2}{c}{AVE}
      & \multicolumn{2}{c}{AVVP}
      & \multicolumn{2}{c}{AVE$\rightarrow$AVVP}
      & \multicolumn{2}{c}{UCF(v)$\leftrightarrow$VGG(a)}
      & \multirow{2}{*}{Avg.} \\
    \cmidrule(lr){2-3}\cmidrule(lr){4-5}\cmidrule(lr){6-7}\cmidrule(lr){8-9}
      & V$\rightarrow$A & A$\rightarrow$V
      & V$\rightarrow$A & A$\rightarrow$V
      & V$\rightarrow$A & A$\rightarrow$V
      & V$\rightarrow$A & A$\rightarrow$V & \\
    \midrule
    \textbf{Ours} (audio semantic anchor)
      & \textbf{57.1} & \textbf{59.4}
      & \textbf{73.4} & \textbf{70.8}
      & \textbf{52.5} & 51.3
      & \textbf{70.3} & \textbf{63.1} & \textbf{62.24} \\
    Symmetric audio-video structure
      & 51.5 & 52.4 & 70.1 & 67.4
      & 51.3 & \textbf{51.6} & 65.4 & 61.5 & 58.90 \\
    Video semantic anchor
      & 54.7 & 54.9 & 69.0 & 66.1
      & 49.7 & 48.1 & 66.5 & 62.3 & 58.91 \\
    \bottomrule
  \end{tabular}}

\end{minipage}

\end{table}

\subsubsection{Performance Analysis}

\autoref{tab:cmg} shows that AHA outperforms Unicode on all eight CMG transfer settings, with the largest gains on AVVP, a localization benchmark that is particularly sensitive to fine-grained temporal cross-modal alignment.

Improvements persist under dataset shift (AVE $\rightarrow$ AVVP and UCF(v) $\leftrightarrow$ VGG(a)), suggesting that AHA generalizes beyond in-distribution evaluation and can transfer discriminative structure learned on AVE to fine-grained localization on AVVP.

Compared with AVT baselines (DCID/FCID) that use extra text supervision as a stronger semantic bridge across datasets, AHA remains competitive using only audio--video pairs and achieves the best results on AVE/AVVP/UCF(v) $\leftrightarrow$ VGG(a), while AVT methods are stronger on AVE $\rightarrow$ AVVP.

Overall, these results suggest that our approach can more effectively extract shared semantics from pure audio--video pairs and support robust cross-modal transfer. 

\subsection{Ablation Study}

Compared with traditional CMG models, AHA differs in three aspects: \textbf{(i)} an asymmetric architecture with an audio semantic anchor; \textbf{(ii)} GRL-based adversarial decoupler instead of MI-estimation-based decoupler; \textbf{(iii)} a local sliding alignment objective for fine-grained temporal alignment. We ablate each component while keeping the CMG protocol and downstream heads unchanged. Since the impact of Cross-CPC and MM-EMA has been discussed in previous work, we do not revisit it here.

\textbf{Local Sliding Alignment (w/o $L_{\text{align}}$)}: As shown in \autoref{tab:loss}, removing $L_{\text{align}}$ consistently degrades performance, with the largest drop on AVVP. This matches our motivation that local sliding alignment provides a tolerant temporal correspondence constraint, sharpening segment-level audio--video matching under small offsets. Notably, the model remains clearly stronger than Unicode even without this term, suggesting that gains are not merely from adding another contrastive loss. Rather, $L_{\text{align}}$ mainly refines fine-grained alignment on top of a transferable backbone.

\textbf{Disentanglement (w/o $L_{\text{grl}}$)}: Removing GRL-based decoupler causes the largest degradation, especially on AVE and AVVP. Under CMG, the classifier is trained on the source modality only, without explicit disentanglement, modality-identifying cues can leak into the modality-specific branches during source training, creating a pronounced train--test mismatch when transferred to the target modality. These results indicate that $L_{\text{grl}}$ is crucial for suppressing modality-specific leakage and enabling reliable zero-shot transfer.

\textbf{Replacing GRL with CLUB ($L_{\text{grl}} \rightarrow L_{\text{CLUB}}$)}: Replacing GRL with CLUB partially recovers the drop but still underperforms the full model, with notable gaps on AVVP. This suggests that GRL-based adversarial decoupler provides a more effective and stable training signal than MI upper-bound minimization.

After validating the contributions of $L_{\text{infonce}}$ and GRL disentanglement, we further ablate the architecture along two axes: \textbf{(i)} symmetric vs.\ asymmetric structure; \textbf{(ii)} the choice of semantic anchor \autoref{tab:structure}.

\textbf{Symmetric vs.\ Asymmetric}: The symmetric dual audio--video structure (Details are shown in \autoref{CMGTasks}) is consistently weaker than our asymmetric design, with clear drops on AVE and UCF(v) $\leftrightarrow$ VGG(a). This is consistent with our analysis of allocation ambiguity, without a directed anchor, semantic information can drift into modality-specific branches, weakening the shared code used for cross-modal transfer.

\textbf{Choice of Anchor}: Within the asymmetric framework, the audio anchor yields the strongest overall transfer and improves most on AVVP, where fine-grained temporal alignment is critical. Under dataset shift (AVE $\rightarrow$ AVVP), the video anchor degrades noticeably, while the audio anchor remains substantially higher, the symmetric structure is comparable on A $\rightarrow$ V but is weaker overall due to its larger drops on AVVP. These results support audio anchoring as a directional constraint that reduces semantic allocation ambiguity and yields a more transferable modality-invariant code in CMG tasks.

\begin{figure}[htbp]
  \centering
  \includegraphics[width=\columnwidth]{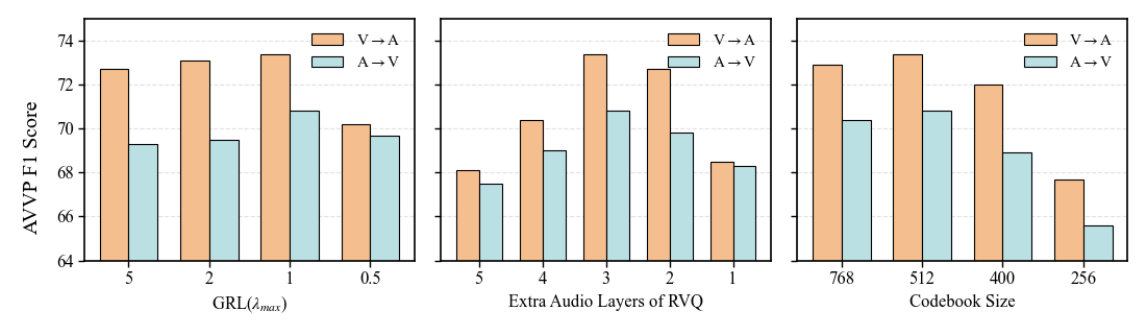}
  \caption{The impact of GRL's adversarial intensity (larger $\lambda_{max}$ indicates stronger adversarial effect), RVQ's number of extra audio layers and Codebook size on experimental accuracy. Additional ablations on the anchor-layer choice $k$ and LSA hyperparameters are provided in \autoref{MoreAblation}.}
  \label{fig:parameter}
\end{figure}

\subsection{Talking Face Disentanglement Experiment}

\begin{figure}[htbp]
  \centering
  \includegraphics[width=\textwidth]{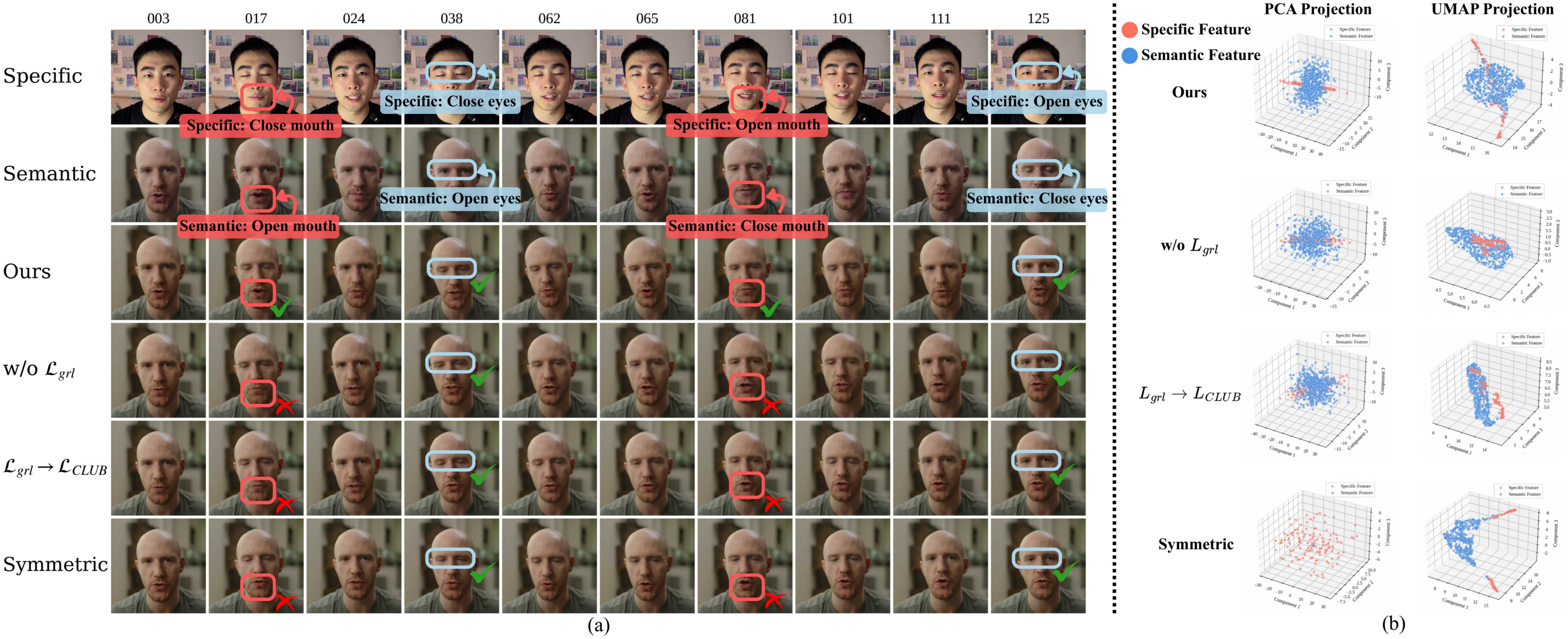}
  \caption{(a) \textbf{Talking Face Disentanglement Experiment Results}: We selected two videos from sources not in the training set to demonstrate the effect. To better showcase the experimental results, we selected frames where Specific Video and Semantic Video have significant differences (for example: whether eyes are closed and mouth is open) for comparison. We set the identity information as the corresponding identity vector from the Semantic Video. For comparison, we also tested (i) removing GRL from our asymmetric structure (w/o $L_{\text{grl}}$). (ii) replacing GRL with CLUB in the asymmetric structure($L_{\text{grl}} \rightarrow L_{\text{CLUB}}$). (iii) adding GRL to a symmetric structure identical to Unicode (Symmetric). (Please see the \autoref{TalkingFaceExperiment} for more experimental results); (b) \textbf{PCA \& UMAP Results}: For the same selected video features, we extracted specific features and semantic features, and performed Principal Component Analysis (PCA) and Uniform Manifold Approximation and Projection (UMAP).}
  \label{fig:face}
\end{figure}

To address the limitation that previous CMG experiments could only implicitly observe semantic information and disentanglement through data, which lacks intuitiveness, we propose a novel Talking Face Disentanglement Experiment that explicitly tests the model's disentanglement capability.

\textbf{Experimental Settings}: Based on our original pipeline~\autoref{fig:pipeline}(a), we use the motion encoder from LIA~\cite{lia, ki2025float} as the video backbone to extract identity information from the input target video that represents its identity, and extracts video features that contain highly entangled motion information of various facial features. In addition, we use Wav2Vec2.0~\cite{wav2vec} as the audio backbone to extract audio features. To better demonstrate the effect of modality transfer, consistent with the CMG experiment approach, we add very sparse phoneme supervision on the time step at the audio shared RVQ side of all models. We selected a filtered TalkVid~\cite{talkvid} subset (approximately 30.6 hours) and trained for approximately ten hours on four A6000 GPUs (Please see the \autoref{TalkingFaceExperiment} for more implementation details).

\textbf{Expected Results}: According to our hypothesis about the Talking Face Disentanglement Experiment, the video information shared with audio should be related to mouth movements, while information representing other facial actions should flow to modality-specific branches. Therefore, the ideal effect of our Talking Face Disentanglement Experiment is to achieve transfer of mouth movements. To elaborate, we can input a semantic video and extract the motion vector representing audio-related mouth movements after passing through the shared RVQ in its pipeline. For another specific video, we can extract the motion vector representing non-mouth movements (such as eye blinking) after passing through the specific encoder in its pipeline. These two can be concatenated and output, then passed through LIA and concatenated with the target identity information, and rendered through LIA's renderer. Finally, a fused video is generated that combines the mouth movements from the semantic video and audio-unrelated movements from the specific video. This also validates the model's decoupling ability, namely whether mixed features common to both videos will appear.

\textbf{Result Analysis}: From \autoref{fig:face}(a), we can see that our current AHA structure can effectively decouple information, enabling alignment of audio-irrelevant actions (such as eye movements) in specific video and alignment of mouth movements in semantic video, with minimal information leakage. When removing the GRL loss (w/o $L_{\text{grl}}$), the disentanglement effect becomes significantly worse, as evidenced by many cases where mouth movements are closed at the same frame in semantic video, while the frame without GRL has the mouth open, and vice versa. There are also cases where eye movements do not correspond to the corresponding frames in specific video. After replacing GRL with CLUB ($L_{\text{grl}} \rightarrow L_{\text{CLUB}}$) and retraining, the improvement in overall disentanglement effect is not obvious. This demonstrates that variational MI estimators are weaker than GRL in handling information disentanglement in such complex information environments. For the symmetric structure, despite strong decoupling with GRL, the lack of audio as an anchor point for semantic information causes most information to leak into the modality-specific branches, resulting in poor disentanglement performance.

\subsubsection{Further Analysis}

\begin{table}[htbp]
\centering
\renewcommand{\arraystretch}{1.0}

\begin{minipage}[t]{0.5\linewidth}
  \centering
  \captionof{table}{Quantitative comparison in Visual-to-Visual Lip
    Synchronization (V2V-LS) and Mouth Landmark RMSE.}
  \label{tab:vsd}
  \vspace{4pt}
  \resizebox{\linewidth}{!}{
  \begin{tabular}{lcccc}
    \toprule
     & Ours & w/o $L_{\text{grl}}$ & $L_{\text{grl}} \rightarrow L_{\text{CLUB}}$ & Symmetric \\
    \midrule
    V2V-LS $\downarrow$     & \textbf{5.98} &  6.77 &  6.82 &  6.40 \\
    Mouth RMSE $\downarrow$ & \textbf{5.51} & 16.59 & 18.26 & 14.61 \\
    \bottomrule
  \end{tabular}}
\end{minipage}%
\hfill
\begin{minipage}[t]{0.46\linewidth}
  \centering
  \captionof{table}{Quantitative comparison in Peak Signal-to-Noise
    Ratio (PSNR) and Learned Perceptual Image Patch Similarity (LPIPS).}
  \label{tab:psnr}
  \vspace{4pt}
  \resizebox{\linewidth}{!}{
  \begin{tabular}{lcccc}
    \toprule
     & Ours & w/o $L_{\text{grl}}$ & $L_{\text{grl}} \rightarrow L_{\text{CLUB}}$ & Symmetric \\
    \midrule
    PSNR $\uparrow$    & \textbf{29.42}  & 26.46  & 27.49  & 26.76  \\
    LPIPS $\downarrow$ & \textbf{0.0468} & 0.0703 & 0.0575 & 0.0730 \\
    \bottomrule
  \end{tabular}}
\end{minipage}

\end{table}

PCA reflects global variance directions. As shown in \autoref{fig:face}, AHA yields clearly different variance patterns with limited overlap between semantic and specific features, suggesting weaker cross-branch leakage. Removing GRL or replacing it with CLUB results in heavier overlap, while the symmetric structure is dominated by the higher-variance branch, consistent with feature collapse rather than structural disentanglement. UMAP further confirms this: under AHA, semantic and specific features show more separated neighborhoods, whereas other variants exhibit highly mixed local structures and similar manifold backbones.

We further evaluate lip synchronization via V2V-LS (frame-wise lip semantic embedding differences~\cite{shi2022learning}) and Mouth RMSE (geometric misalignment of 2D mouth landmarks), and reconstruction quality via PSNR and LPIPS. As shown in \autoref{tab:vsd} and \autoref{tab:psnr}, AHA achieves the best performance across all four metrics, indicating improved semantic coherence, reconstruction fidelity, and perceptual quality. Detailed metric definitions are provided in \autoref{TalkingFaceExperiment}.




\section{Conclusion}

This paper investigates audio-visual joint representation learning under Cross-Modal Generalization (CMG), noting that current symmetric discrete unified representation frameworks are prone to information allocation ambiguity. We propose Asymmetric Hierarchical Anchoring (AHA), which leverages the hierarchical structure of audio RVQ to construct shared semantic anchors for directed video-to-audio semantic distillation. AHA further introduces a GRL-based adversarial decoupler to suppress semantic information leakage and Local Sliding Alignment (LSA) to improve fine-grained temporal alignment. Extensive experiments on AVE, AVVP, cross-dataset transfer, and a Talking Face Disentanglement Experiment show that AHA improves cross-modal generalization, disentanglement quality, and reconstruction fidelity.

\textbf{Limitations}:
While AHA demonstrates strong performance under the current CMG setting, this work mainly focuses on audio-visual representation learning with a fixed semantic anchor and standard downstream evaluation protocols. Future work will explore adaptive anchor selection across tasks and modality pairs, extend AHA to broader multimodal settings, and conduct more systematic evaluations under open-domain noise, weak cross-modal correspondence, and long temporal contexts.

\bibliography{example_paper}
\bibliographystyle{nips}

\newpage
\appendix



\section{Implementation of Adversarial Disentanglement(Section~\ref{adv_grl})}
\label{alg:grl_velocity}

\begin{algorithm}[th]
\caption{Adversarial Disentanglement with Velocity-Aware Sampling}
\begin{algorithmic}

\REQUIRE Video Encoders $E_{v\_sem}, E_{v\_spec}$; Discriminator $D_{\phi}$; Batch $\mathcal{B}$.

 Hyperparameters: Max steps $N_{max}$; Temp $\tau$; Smoothing $\epsilon$; GRL Limit $\lambda_{max}$; Learning rate $\alpha$; Model parameters $\theta$.
 
 Helper Functions: Score $s(\mathbf{x}, \mathbf{y}) = D_\phi(\mathcal{R}_\lambda(\mathbf{x}), \mathbf{y})$.
\ENSURE Disentangled Specific Features $\mathbf{z}_{spec}$.

\STATE Initialize step $p \leftarrow 0$
\WHILE{$p < N_{max}$}
    \STATE \textcolor{blue}{\textbf{1. Feature Extraction}}
    \STATE $V \leftarrow \text{Quantizer}(E_{v\_sem}(\mathcal{B}))$;  \COMMENT{Video Units}
    \STATE $v \leftarrow E_{v\_spec}(\mathcal{B})$  \COMMENT{Video Specific Features}

    \STATE \textcolor{blue}{\textbf{2. Velocity-Aware Sampling}}
    \FOR{each sequence in batch}
        \STATE $\delta_t \leftarrow \| V_{t+1} - V_{t} \|_2$ 
        \STATE $P(t) \leftarrow \delta_t / (\sum \delta_\tau + \epsilon)$; 
        \STATE Sample $\mathcal{K} \sim \text{Multinomial}(P(t))$
        \STATE $\hat{V}, \hat{v} \leftarrow \text{Gather}(\{V, v\}, \mathcal{K})$
    \ENDFOR

    \STATE \textcolor{blue}{\textbf{3. Adversarial Objective}}
    \STATE $\lambda \leftarrow \lambda_{max} \cdot \left( \frac{2}{1 + \exp(-10 \cdot p/N_{max})} - 1 \right)$
    \STATE Sample negatives $\mathcal{N}$ from batch
    \STATE $\mathcal{L}_{grl} \leftarrow -\log \frac{\exp(s(\hat{v}, \hat{V})/\tau)}{\sum_{V' \in \{\hat{V}\} \cup \mathcal{N}} \exp(s(\hat{v}, V')/\tau)}$

    \STATE \textcolor{blue}{\textbf{4. Optimization}}
    \STATE \COMMENT{Update Discriminator (Minimize Loss)}
    \STATE $\theta_D \leftarrow \theta_D - \alpha \nabla_{\theta_D} \mathcal{L}_{grl}$
    
    \STATE \COMMENT{Update Specific Encoder (Maximize Loss via GRL)}
    \STATE $\theta_{E_{v\_spec}} \leftarrow \theta_{E_{v\_spec}} - \alpha \lambda \nabla_{\theta_{E_{v\_spec}}} \mathcal{L}_{grl}$
    \STATE \COMMENT{gradient is negated by $\mathcal{R}_\lambda$ during backprop, 
 achieving effective gradient ascent on $E_{v\_spec}$}
    \STATE $p \leftarrow p + 1$
\ENDWHILE
\end{algorithmic}
\end{algorithm}

\section{Details of CMG Downstream Tasks}
\label{CMGTasks}

\textbf{Pre-train Details:}

Following the standard CMG evaluation protocol, we adopt the same backbone architectures as Unicode~\cite{unicode}. Specifically, for each 1-second visual segment we uniformly sample 16 RGB frames and extract deep convolutional feature maps from VGG-19 after the final pooling stage; we then apply global average pooling across the 16 frames to obtain $7 \times 7 \times 512$-D visual feature maps. For the audio stream, we employ a VGG-like network pre-trained on AudioSet to extract 128-D features for each 1-second audio segment. For both modalities, we further project semantic representations into a shared 512-D embedding space.

To capture modality-specific information, the video branch uses a convolutional encoder to produce spatial feature maps of shape $3 \times 3 \times 2048$; we apply spatial average pooling followed by a linear projection to obtain video-specific features of shape $T \times 512$. The audio branch uses a linear encoder to produce audio features of shape $T \times 256$. The video-specific features are used for reconstruction and for adversarial disentanglement via the GRL-based decoupler (operating against video semantic units). The audio and video semantic units are aligned via Local Sliding Alignment. In contrast to prior unified-representation approaches (e.g., DCID~\cite{fcid}, FCID~\cite{fcid}) that rely on additional text supervision, our Asymmetric Hierarchical Anchoring (AHA) framework is pre-trained solely on paired audio--video data in this downstream tasks.

For Cross-CPC, we set the prediction horizon to one step and apply it between the two modality semantic sequences during unified-representation pre-training. For GRL-based disentanglement, we first update the GRL discriminators once with frozen encoders, and then update the full model (encoder, CPC, decoder) with GRL reversal in the main pass. We use a learning rate of $1\times10^{-4}$, set $\gamma=0.99$ for MM-EMA, and adopt a batch size of 96.

\textbf{Details of all downstream tasks:}

\begin{table}[htbp]
\centering
\footnotesize
\setlength{\tabcolsep}{12pt}
\renewcommand{\arraystretch}{1.0}
\caption{Details of all downstream tasks.}
\label{tab:downstream_tasks}
\begin{tabular}{l c c c}
\toprule
\textbf{Task} & \textbf{Pretrained Modality} & \textbf{Downstream Dataset} & \textbf{Generalization Direction} \\
\midrule
\multirow{2}{*}{\begin{tabular}[c]{@{}l@{}}Cross-Modal\\Event Classification\end{tabular}}
& \multirow{2}{*}{Audio-Visual}      & \multirow{2}{*}{AVE} & A$\rightarrow$V \\
&  &                         & V$\rightarrow$A \\
\midrule
\multirow{2}{*}{\begin{tabular}[c]{@{}l@{}}Cross-Modal\\Event Localization\end{tabular}}
& \multirow{2}{*}{Audio-Visual}      & \multirow{2}{*}{AVVP} & A$\rightarrow$V \\
&  &                        & V$\rightarrow$A \\
\midrule
\multirow{3}{*}{\begin{tabular}[c]{@{}l@{}}Cross Modal \& Dataset\\Event Localization\end{tabular}}
& \multirow{3}{*}{Audio-Visual}      & \multirow{3}{*}{AVE \& AVVP} & A(AVE)$\rightarrow$V(AVVP) \\
&  &                               & V(AVE)$\rightarrow$A(AVVP) \\
&  &                               & UCF(v)$\leftrightarrow$VGG(a) \\
\bottomrule
\end{tabular}
\end{table}

\textbf{Cross-Modal Event Classification (AVE)}~\cite{AVE}: The AVE dataset contains 28 event categories and each audio/video clip is 10 seconds long. Given discrete representations extracted from one modality, we attach a two-layer MLP classifier to map the sequence into 28-way class logits, followed by a softmax and a standard cross-entropy loss. To test video-to-audio generalization (V$\rightarrow$A), we train the classifier using video-derived discrete sequences and then directly replace the input with audio-derived sequences at test time, keeping both the frozen encoder and the learned classifier unchanged; audio-to-video (A$\rightarrow$V) is evaluated analogously.

\textbf{Cross-Modal Event Localization (AVVP)}~\cite{AVVP}: The AVVP dataset includes 25 event categories, and a single clip may contain multiple events. We again use the frozen encoder to obtain a length-$T$ discrete sequence, and employ a two-layer MLP localization head to predict 25-dimensional event scores (per segment). We apply a sigmoid activation and optimize with a multi-label classification objective (binary cross-entropy) against the ground-truth labels. All remaining training settings follow those used for cross-modal event classification.

\textbf{Cross-Modal \& Cross-Dataset Event Localization (AVE $\rightarrow$ AVVP; UCF(v)$\leftrightarrow$VGG(a))}: To further assess transferability across datasets, we consider the 12 overlapping event categories shared by AVE and AVVP: dog, car, helicopter, violin fiddle, frying food, motorcycle, acoustic guitar, banjo, baby cry, chainsaw, cat, accordion. We train the localization head using single-modality inputs from the AVE training set and directly evaluate on the opposite modality in the AVVP validation set, without any additional adaptation. Following prior practice, we report the F1 score as the primary metric, while keeping the optimization protocol identical to the AVVP localization setting. We also tested cross-modal classification tasks, performing classification between the visual modality on a subset of UCF-101~\cite{soomro2012ucf101} (16 classes) and the audio modality on a subset of VGGSound-AVEL~\cite{vggsound2,vggsound1} (16 classes).

\begin{figure*}[htbp]
  \centering
  \includegraphics[width=0.8\textwidth]{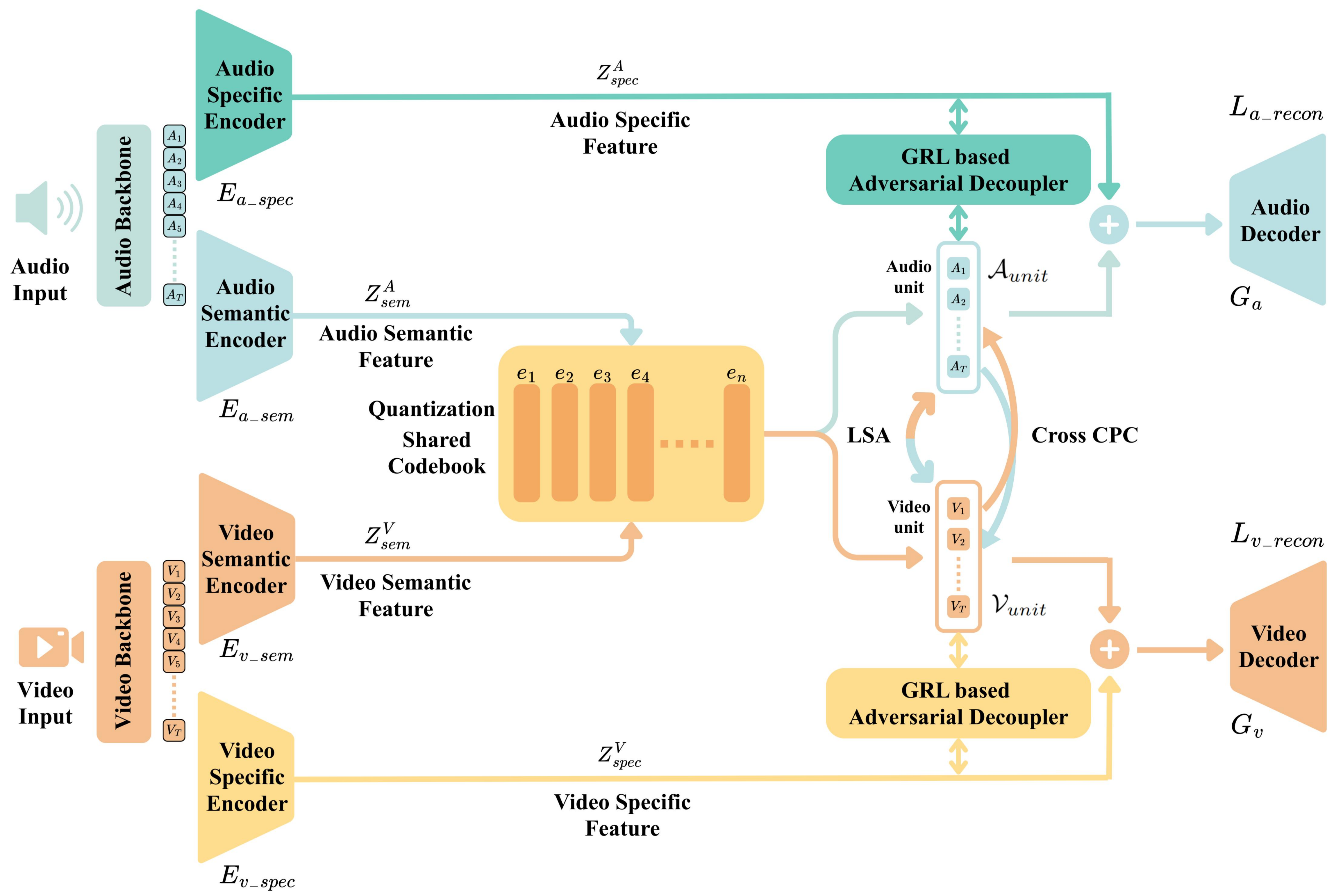}
  \caption{Brief Pipeline of symmetric structure}
  \label{fig:pipelinesymmetric}
\end{figure*}

\textbf{Brief Pipeline of symmetric structure}: Below is a schematic diagram of the symmetric structure in CMG Downstream Tasks and in the Talking Face Disentanglement Experiment. Except for slight differences in the peripheral structure (i.e., audio or video backbone, encoder, decoder), the main structure is shown in \autoref{fig:pipelinesymmetric}.

\begin{table}[h]
\centering
\caption{Extension to AVT setting.}
\label{tab:avt}
\resizebox{\linewidth}{!}{%
\begin{tabular}{lcccccccccc}
\toprule
Method & \multicolumn{2}{c}{AVE} & \multicolumn{2}{c}{AVVP}
       & \multicolumn{2}{c}{AVE$\to$AVVP} & \multicolumn{2}{c}{UCF$\leftrightarrow$VGG} & Avg. \\
       & V$\to$A & A$\to$V & V$\to$A & A$\to$V
       & V$\to$A & A$\to$V & V$\to$A & A$\to$V & \\
\midrule
DCID (AVT) & 54.5 & 55.0 & 40.9 & 41.6 & 56.5 & 53.6 & 68.1 & 61.7 & 53.99 \\
FCID (AVT) & 55.9 & 55.0 & 43.6 & 45.1 & 57.4 & 58.5 & 69.6 & 62.0 & 55.89 \\
Ours (AVT) & \textbf{59.1} & \textbf{61.7} & \textbf{73.9} & \textbf{72.1}
           & \textbf{59.5} & \textbf{60.3} & \textbf{73.5} & \textbf{66.8} & \textbf{65.86} \\
\bottomrule
\end{tabular}}
\end{table}

\textbf{Extension to Text--Vision--Audio (AVT) Setting}: Following DCID/FCID~\cite{fcid}, we incorporate BERT-based text
features while keeping audio as the semantic anchor.
As shown in~\autoref{tab:avt}, AHA (AVT) achieves the best results across all metrics,
confirming extensibility beyond the pure audio--visual setting.

\section{Details of Talking Face Disentanglement Experiment}
\label{TalkingFaceExperiment}

\begin{figure*}[htbp]
  \centering
  \includegraphics[width=\textwidth]{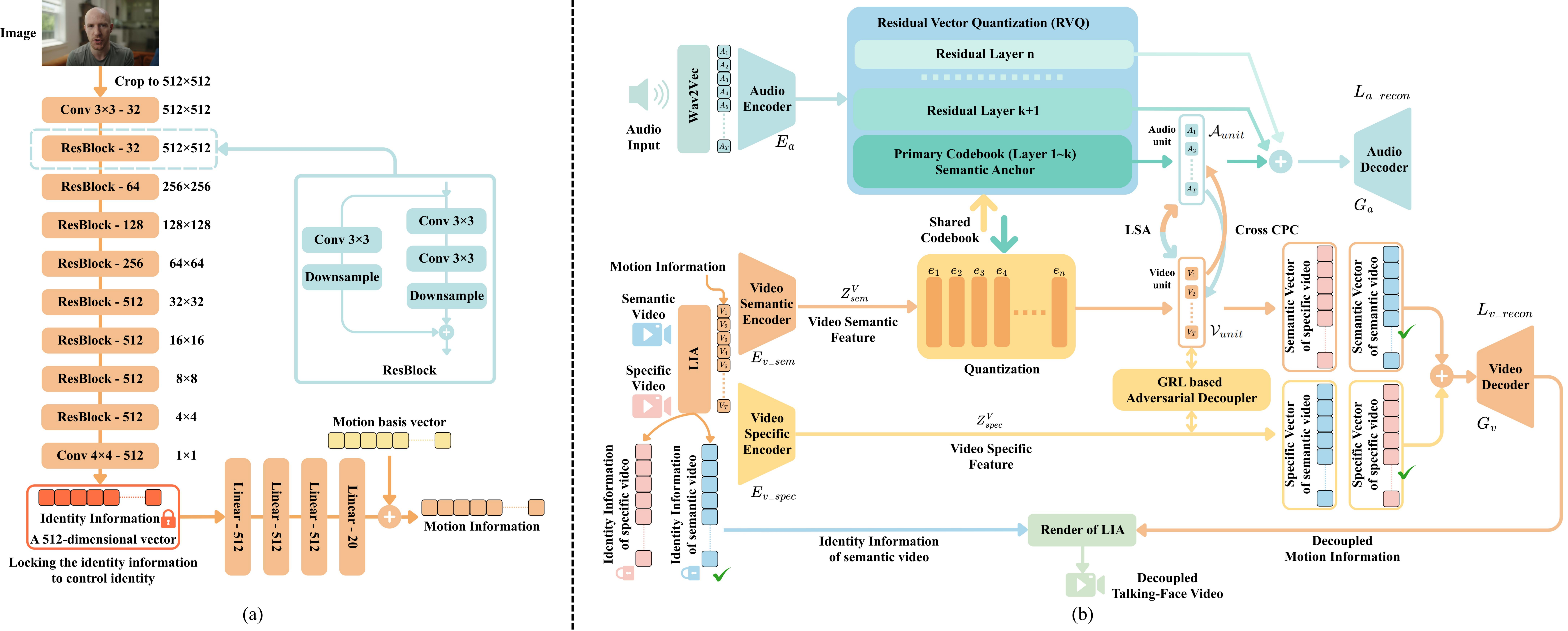}
  \caption{(a) LIA's Brief Pipeline for Extracting Identity Information and Motion Information; (b) Brief Pipeline for Talking Face Disentanglement Experiment}
  \label{fig:pipelineface}
\end{figure*}

Based on our initial AHA Pipeline, we replaced the Video backbone with LIA~\cite{lia,ki2025float} and the Audio Backbone with Wav2Vec2.0~\cite{wav2vec}. The pipeline for the Talking-Face Disentanglement Experiment is shown in \autoref{fig:pipelineface}(b). For the Video Backbone, LIA's encoder extracts Identity Information from the input target image that represents identity of
this image, and extracts Motion Information containing highly entangled facial features from it. We use this Motion Information as the video feature, with the detailed pipeline shown in \autoref{fig:pipelineface}(a). Additionally, for the Audio Backbone, we used Wav2Vec to extract audio features. After processing with LIA and Wav2Vec, the final video input dimension and audio input dimension are both 512-D.

Specifically, for our decoupling process, we select a video not present in the training data as the Semantic Video. After passing through the Encoder of LIA, we obtain the Identity Information of Semantic Video and Motion Information of Semantic Video. Additionally, we select another video also not present in the training data as the Specific Video. After similarly passing through the Encoder of LIA, we obtain the Identity Information of Specific Video and Motion Information of Specific Video. We freeze the obtained Identity Information and input the Motion Information into our AHA, obtaining the Semantic Vector of Semantic Video, Semantic Vector of Specific Video, Specific Vector of Semantic Video, and Specific Vector of Specific Video respectively. Finally, we select the Semantic Vector of Semantic Video and Specific Vector of Specific Video, concatenate them, and send them to the Decoder to obtain the Decoupled Motion Information. Together with the frozen Identity Information of Semantic Video, we input them into the Renderer of LIA to finally obtain our Decoupled Talking-Face Video.

Therefore, the expected effect is to generate a fused Decoupled Talking-Face Video that combines the mouth movements obtained from the Semantic Video with the audio-independent motions (such as blinking actions) from the Specific Video.

For our training details, we selected a subset of the TalkVid~\cite{talkvid} dataset (approximately 38.4 hours). After our data processing, we ultimately used approximately 30.6 hours of data and trained for approximately 10 hours across 4 A6000 GPUs, covering 250 epochs.

\textbf{Data Processing Details:}

\begin{enumerate}
    \item Face Recognition: Due to transitions or narration in YouTube videos, TalkVid may contain video segments with audio but no actual human faces. Therefore, we used Face Recognition to detect whether the video contains human faces and removed videos without faces.
    \item Silence Detection: Since some videos contain prolonged silence (exceeding 1/3 of the video duration), which is useless for training purposes, we also performed silence detection on the videos.
    \item Audio-Visual Synchronization: Since most data in TalkVid comes from YouTube, some videos are similar to film reviews or commentary videos that contain facial action videos, but the audio and corresponding video cannot be properly synchronized. Based on this problem, we made certain adjustments based on the Python code from~\cite{chung2016out}, and filtered out videos that do not meet our requirements for audio-visual synchronization.
    \item Phoneme Extraction: To better demonstrate the modality transfer effect, consistent with the CMG experiment approach, we add a phoneme supervision on the audio shared RVQ that is very sparse in the time dimension (statistically around 20\% of frames participate), i.e., adding a simple classification head to perform phoneme classification on labeled frames. We use BFA~\cite{rehman2025bfa} to extract phonemes.
    \item Large Angle Face Detection: Since some videos contain significant angle changes in face orientation, mouth information may be lost due to large-angle movements, which affects our experiments. Therefore, we used~\cite{hempel20226d} to detect large angles in videos and removed videos where face rotation exceeds 60 degrees from the dataset.
\end{enumerate}

\textbf{More Talking Face Disentanglement Experiment Examples:}

\begin{figure}[htbp]
  \centering
  \includegraphics[width=\textwidth]{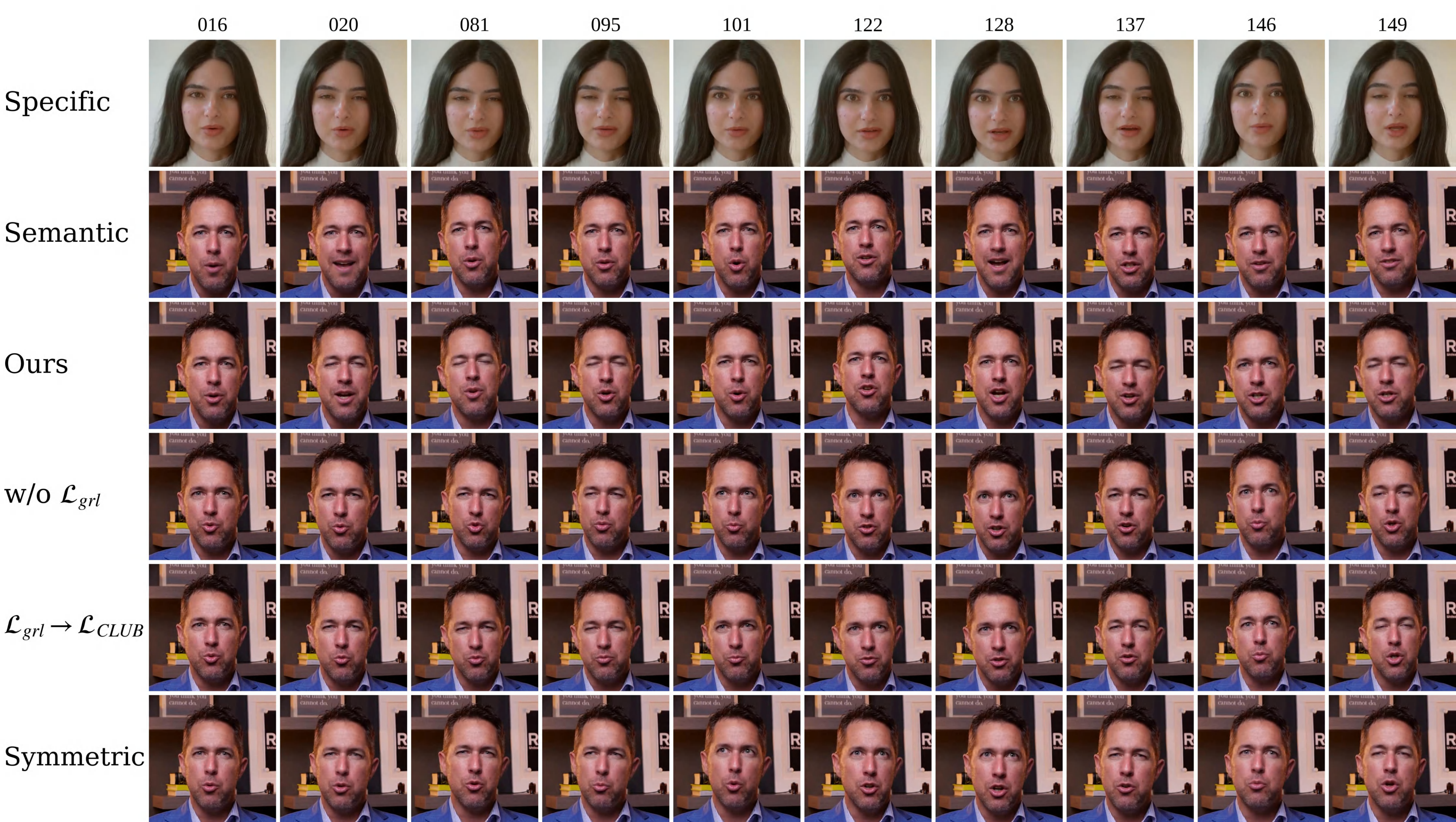}
  \caption{More Talking Face Disentanglement Experiment Examples (1) }
  \label{fig:compare1}
\end{figure}

\begin{figure}[htbp]
  \centering
  \includegraphics[width=\textwidth]{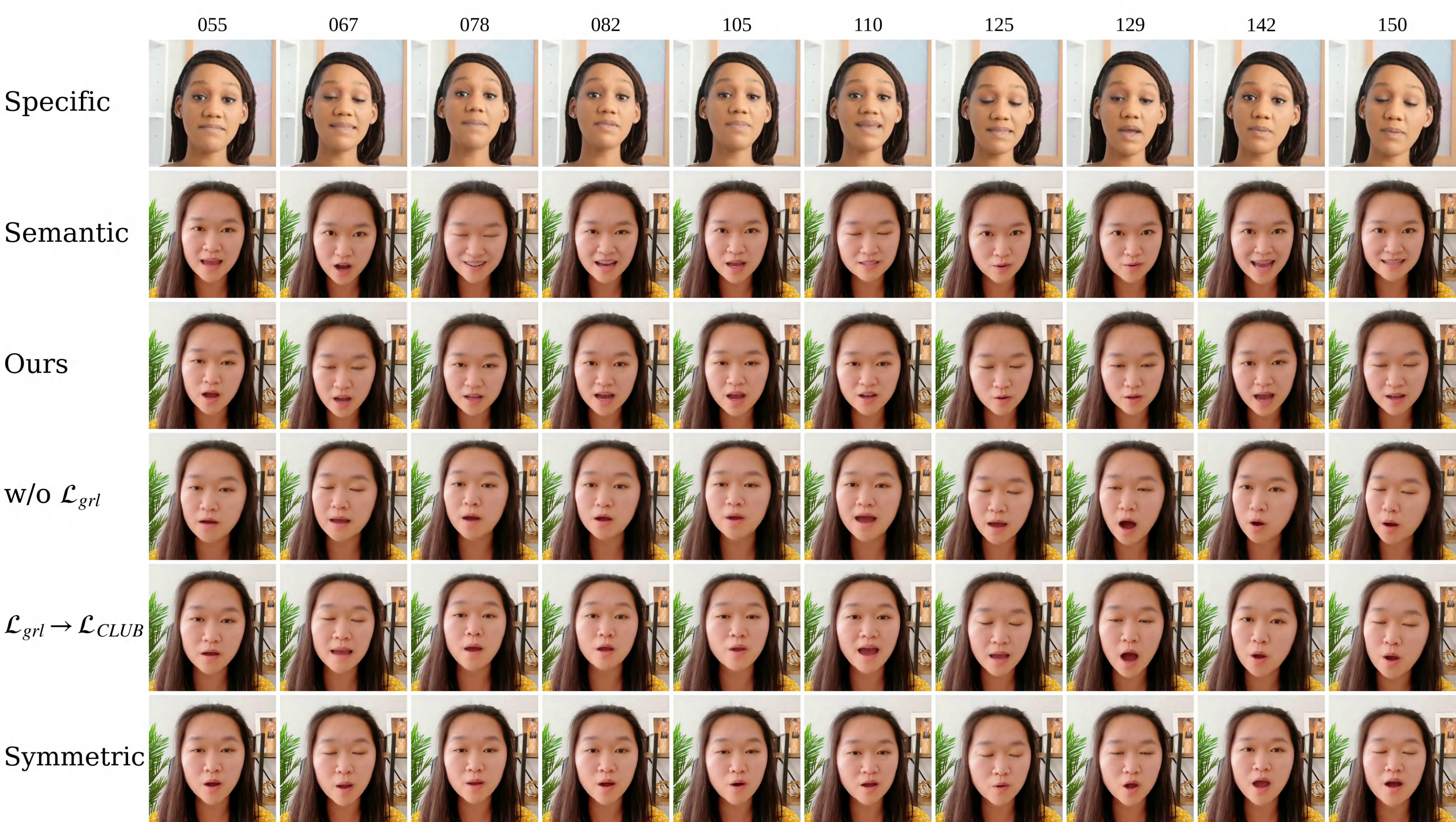}
  \caption{More Talking Face Disentanglement Experiment Examples (2) }
  \label{fig:compare2}
\end{figure}

More experimental results can be found in \autoref{fig:compare1} and \autoref{fig:compare2}. From the extensive experimental results, we can see the robustness of our model in terms of decoupling stability and reconstruction fidelity.

\textbf{Detailed diagrams of PCA and UMAP:}

\begin{figure*}[h]
  \centering
  \includegraphics[width=\textwidth]{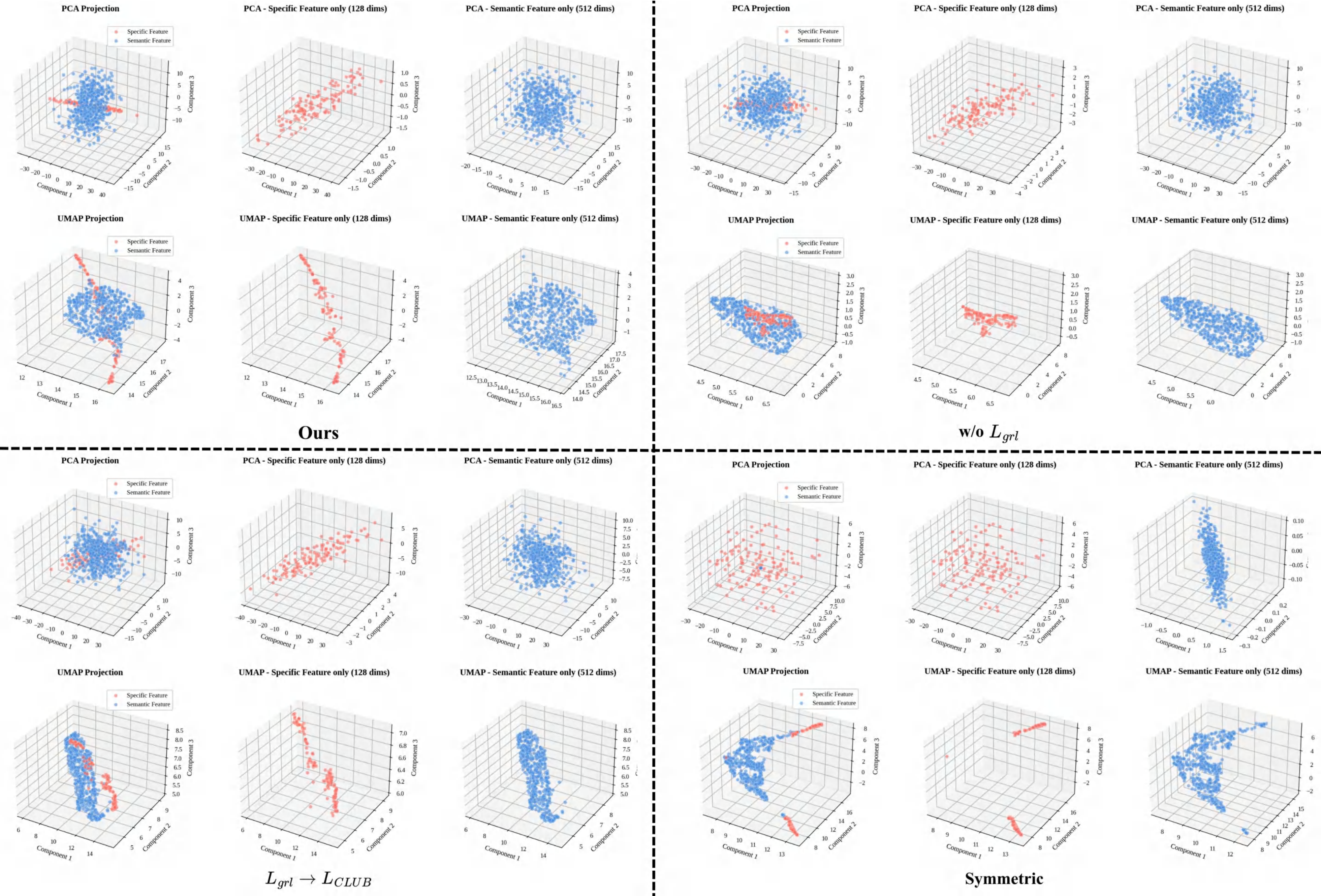}
  \caption{Detailed diagrams of PCA and UMAP}
  \label{fig:pcaumap}
\end{figure*}

To more clearly see the individual distribution of specific features and semantic features respectively, we present the PCA and UMAP details of specific feature only and semantic feature only in \autoref{fig:pcaumap}.

\textbf{Evaluation Metrics:}

\begin{enumerate}
    \item \textbf{PSNR}~\cite{5596999}: Peak Signal-to-Noise Ratio (PSNR) measures the reconstruction quality of an image by comparing it with the reference image, higher values indicate better fidelity.
 
    \item \textbf{LPIPS}~\cite{zhang2018unreasonable}: Learned Perceptual Image Patch Similarity (LPIPS) is used to measure the perceptual similarity between reconstructed image and real image based on the pre-trained AlexNet features.

    \item \textbf{V2V-LS (Ours)}: We propose a video-to-video lip-sync metric inspired by Wav2Lip~\cite{prajwal2020lip}, using visual embeddings from AV-HuBERT~\cite{shi2022learning}. The lip region embeddings of the two videos are compared via frame-wise differences to compute a score, quantifying how well the source video drives the target lip movements.

    \item \textbf{Mouth RMSE}: Root-mean-square error (RMSE) computed over the 2D mouth landmarks of two videos, capturing frame-wise geometric misalignment of the lips; lower values indicate better lip synchronization.

\end{enumerate}

\section{More Ablation Studies}
\label{MoreAblation}
\paragraph{Sensitivity to Anchor Layer $k$.}
~\autoref{tab:k_sensitivity} sweeps $k \in \{1,2,3,4\}$.
Performance is robust across all values; $k{=}1$ and $k{=}2$ yield the best results,
consistent with the hypothesis that primary semantic content concentrates in lower RVQ
layers. We use $k{=}1$ for CMG (fair comparison with single-codebook baselines)
and $k{=}2$ for Talking Face (higher semantic density required).

\begin{table}[htbp]
\centering
\caption{Sensitivity to shared RVQ anchor layers $k$.}
\label{tab:k_sensitivity}
\resizebox{\linewidth}{!}{%
\begin{tabular}{lcccccc}
\toprule
$k$ & AVE V$\to$A & AVE A$\to$V & AVVP V$\to$A & AVVP A$\to$V
    & V2V-LS $\downarrow$ & PSNR $\uparrow$ \\
\midrule
1        & 57.1 & \textbf{59.4}          & 73.4          & 70.8          & 6.43          & 28.01 \\
2        & \textbf{57.6}          & 58.9 & \textbf{73.5} & \textbf{71.4} & \textbf{5.98} & \textbf{29.42} \\
3        & 56.7          & 58.2          & 72.9          & 71.3          & 6.14          & 28.72 \\
4        & 56.4          & 58.3          & 71.7          & 70.2          & 6.77          & 27.65 \\
\bottomrule
\end{tabular}}
\end{table}

\paragraph{Sensitivity to LSA Hyperparameters.}~\autoref{tab:lsa_R}--\autoref{tab:lsa_npos} sweep $R$ and $n_{\text{pos}}$.
Performance plateaus for $R \geq 5$; only $R{=}1$ (overly narrow scope) hurts.
$n_{\text{pos}}{=}1$ is optimal: larger tolerance blurs the alignment target.
Both parameters follow simple task-granularity guidelines and require no per-dataset tuning.

\begin{table}[htbp]
\centering
\caption{LSA search scope $R$ sweep (fixed $n_{\text{pos}}{=}1$).}
\label{tab:lsa_R}
\begin{tabular}{lcccc}
\toprule
$R$ & AVE V$\to$A & AVE A$\to$V & AVVP V$\to$A & AVVP A$\to$V \\
\midrule
1        & 52.6 & 54.7 & 66.7 & 67.9 \\
3        & 55.9 & 58.3 & 72.1 & 69.2 \\
5 (ours) & 57.1 & \textbf{59.4} & 73.4 & \textbf{70.8} \\
7        & \textbf{57.2} & 59.2 & \textbf{73.6} & 70.6 \\
\bottomrule
\end{tabular}
\end{table}

\begin{table}[htbp]
\centering
\caption{LSA positive tolerance $n_{\text{pos}}$ sweep (fixed $R{=}5$).}
\label{tab:lsa_npos}
\begin{tabular}{lcccc}
\toprule
$n_{\text{pos}}$ & AVE V$\to$A & AVE A$\to$V & AVVP V$\to$A & AVVP A$\to$V \\
\midrule
1 (ours) & \textbf{57.1} & \textbf{59.4} & \textbf{73.4} & \textbf{70.8} \\
2        & 56.1 & 57.8 & 70.5 & 68.4 \\
3        & 53.1 & 54.9 & 65.2 & 66.7 \\
\bottomrule
\end{tabular}
\end{table}

\section{Training Details}
\label{train_details}

\begin{figure}[htbp]
  \centering
  \includegraphics[width=0.8\textwidth]{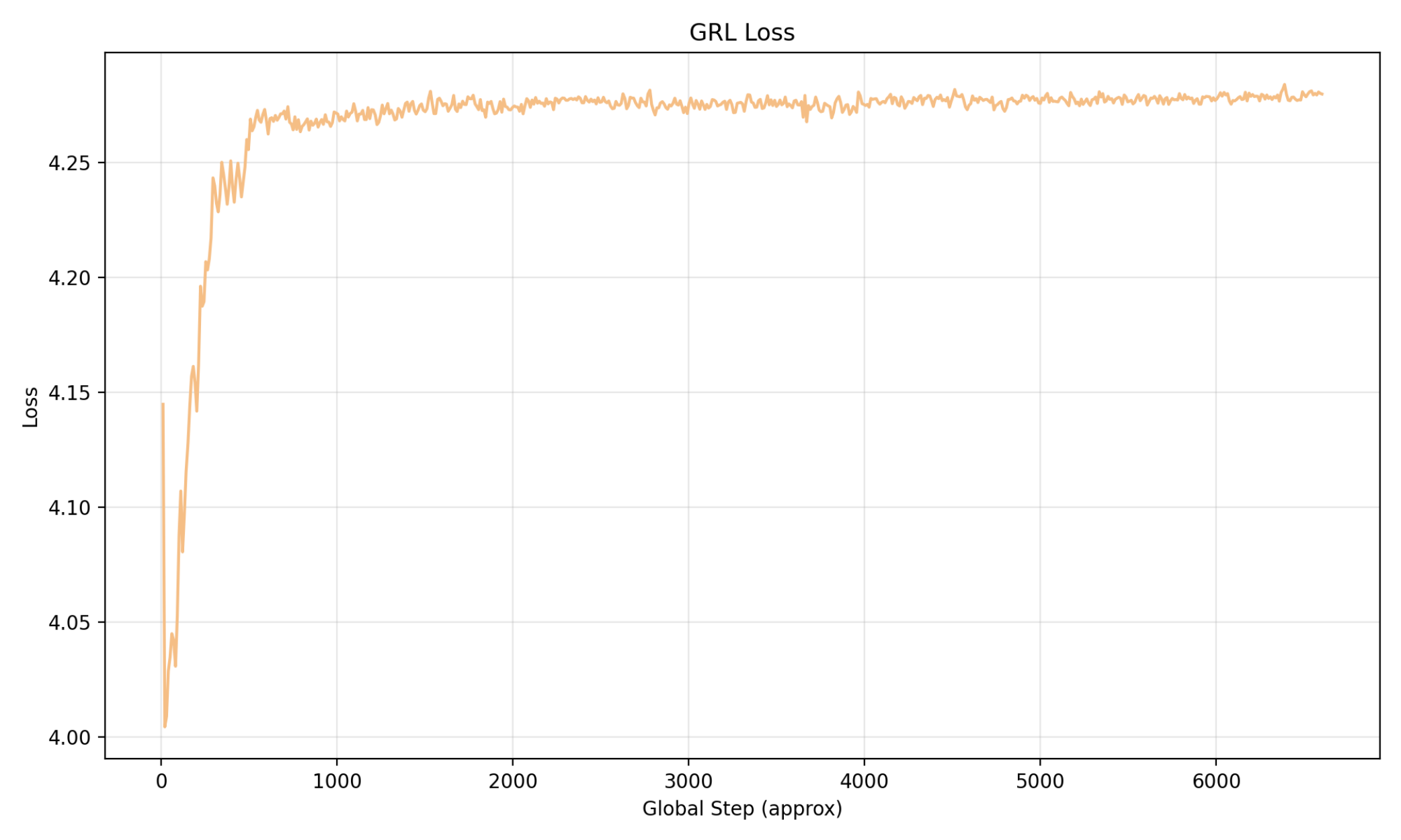}
  \caption{Changes in GRL Loss During Training}
  \label{fig:grl}
\end{figure}

\autoref{tab:hyper} shows a portion of the hyperparameters used during our training process and their corresponding values. \autoref{fig:grl} shows the GRL loss during our training process. As can be seen, after a certain number of steps, the GRL loss stabilizes and fluctuates around 4.28. This value is determined by $ln(batch size \times GRL\ negative\ sample\ ratio) = ln(96\times0.75)= 4.28$ in our experiment. Fluctuations around this value indicate that the GRL adversarial process is proceeding normally during our model training.

\textbf{Training Efficiency}: All models were trained for 250 epochs on 4 NVIDIA A6000 GPUs with batch size 96. Our proposed AHA completed training in approximately \textbf{9.7} hours, while symmetric structure required \textbf{13.3} hours, demonstrating a \textbf{27\%} improvement in training efficiency.


\begin{table}[htbp]
\centering
\footnotesize
\setlength{\tabcolsep}{12pt}
\renewcommand{\arraystretch}{1.0}
\caption{Some hyperparameters and their values.}
\label{tab:hyper}
\begin{tabular}{cc}
\toprule
\textbf{Hyperparameter} & \textbf{Value} \\
\midrule
Codebook size & 512 \\
Embedding dimension & 512 \\
CPC hidden dimension & 256 \\
CPC context dimension & 256 \\
CPC LSTM layers & 2 \\
CPC prediction steps & 1 \\
GRL velocity sample ratio & 0.4 \\
GRL negative sample ratio & 0.75 \\
Main optimizer & AdamW \\
GRL optimizer & Adam (learning rate = 1e-4) \\
Learning rate scheduler & CosineAnnealingWarmupLR \\
Warmup epochs & 5 \\
Minimum learning rate & 1e-6 \\
GRL temperature ratio $\tau$ & 0.1 \\
GRL $\lambda_{max}$ & 1 \\
Audio extra layers & 3 \\
Positive tolerance set (CMG Tasks) & 1 \\
Positive tolerance set (Talking Face Experiment) & 3 \\
LSA window size (CMG Tasks) & 5 \\
LSA window size (Talking Face Experiment) & 31 \\
Shared RVQ Layers (CMG Tasks) & 1 \\
Shared RVQ Layers (Talking Face Experiment) & 2 \\
\bottomrule
\end{tabular}
\end{table}

\section{Existing Assets and Licenses}

We use existing datasets, pretrained components, and reference implementations only for non-commercial academic research and evaluation. We cite the original papers and official sources where applicable, and follow the corresponding licenses or terms of use. The main assets used in this work are summarized below.

\begin{itemize}
    \item \textbf{AVE}~\cite{AVE}: Used for cross-modal event classification and AVE-to-AVVP transfer evaluation. The official release provides dataset access and citation information, but we did not find an explicit public dataset license. We therefore use it only for academic evaluation and cite the original paper.

    \item \textbf{AVVP / LLP}~\cite{AVVP}: Used for cross-modal event localization. The official project repository is released under the GNU General Public License v3.0. We use the released annotations/features only for academic evaluation and cite the original paper.

    \item \textbf{VGGSound and VGGSound-AVEL}~\cite{vggsound2,vggsound1}: Used for audio--visual pre-training and cross-dataset evaluation. VGGSound is released under the Creative Commons Attribution 4.0 International License (CC BY 4.0), while the copyright of the original videos remains with their respective owners.

    \item \textbf{UCF-101}~\cite{soomro2012ucf101}: Used for cross-dataset visual-to-audio evaluation. The official website provides the dataset for research use, but we did not find an explicit public license in the official release. We use it only for academic evaluation and cite the original dataset paper.

    \item \textbf{TalkVid}~\cite{talkvid}: Used in the Talking Face Disentanglement Experiment. The official repository states that the dataset is released under CC BY-NC 4.0 for non-commercial research use, and that the accompanying code is released under the Apache License 2.0.

    \item \textbf{LIA}~\cite{lia}: Used as the motion encoder/renderer in the diagnostic talking-face disentanglement evaluation. The official repository is released under the Creative Commons Attribution-NonCommercial 4.0 International Public License (CC BY-NC 4.0). We use it only for non-commercial research.

    \item \textbf{Wav2Vec2.0 / fairseq}~\cite{wav2vec}: Used as the audio backbone in the talking-face experiment. The fairseq repository is MIT-licensed and states that the license applies to pretrained models as well.

    \item \textbf{VGG-19 and AudioSet-pretrained audio features}: For CMG experiments, we follow the standard protocol and use VGG-19 visual features and AudioSet-pretrained audio features. TorchVision is released under the BSD 3-Clause License, and AudioSet annotations are released under CC BY 4.0, with the AudioSet ontology under CC BY-SA 4.0.

    \item \textbf{BERT-based text features}~\cite{devlin2019bert}: Reference AVT baselines use BERT-based text features. The official Google BERT repository states that its code and pretrained models are released under the Apache License 2.0.
\end{itemize}

For datasets derived from online videos, we use only the released annotations, features, or clips for academic evaluation and respect the terms of the original dataset providers and source platforms.

\section{Broader Impacts}

This work studies audio--visual joint representation learning under Cross-Modal Generalization (CMG). Its potential positive impact lies in improving the robustness of audio--visual representation learning under modality shift and reducing reliance on extensive labeled data, which may benefit downstream audio--visual understanding, localization, and transfer-learning applications.

This work also includes a Talking Face Disentanglement Experiment as a diagnostic evaluation of semantic--specific disentanglement. In this experiment, we use an existing renderer to visualize fused facial motions by combining mouth-related motion from one sequence with other facial motions from another sequence. Although this setting is used only for representation analysis rather than for releasing a deployable face-generation system, related techniques could potentially be misused for synthetic media manipulation, impersonation, or non-consensual deepfakes. The societal risks of deepfakes have been discussed in prior work~\cite{mirsky2021creation}.

We therefore recommend responsible use of this line of research, including using data with appropriate consent and governance, clearly disclosing or labeling synthetic outputs, and conducting safety evaluations before any deployment in controllable audio--visual generation systems. We strongly condemn malicious use and advocate responsible and ethical research practices. If the paper is accepted, we plan to release the code to support transparency and verification of our findings.


\newpage

\end{document}